%% file: Main.tex
\documentclass{article}
\usepackage{ijcai21}

\usepackage{times}
\usepackage{soul}
\usepackage{url}
\usepackage[hidelinks]{hyperref}
\usepackage[utf8]{inputenc}
\usepackage[small]{caption}
\usepackage{graphicx}
\usepackage{amsmath}
\usepackage{amsthm}
\usepackage{booktabs}
\usepackage{algorithm}
\usepackage{algorithmic}
\usepackage{multirow}
\usepackage{amssymb}
\usepackage{color}
\title{Scribble-Supervised Semantic Segmentation by Uncertainty Reduction on Neural Representation and Self-Supervision on Neural Eigenspace}

\author{
Zhiyi Pan$^1$
\and
Peng Jiang$^1$\thanks{Corresponding Author}\and
Yunhai Wang$^1$\and
Changhe Tu$^{1,2}$\and
Anthony G. Cohn$^3$
\affiliations
$^1$Shandong University\\
$^2$AICFVE, Beijing Film Academy\\
$^3$University of Leeds
\emails
\{panzhiyi1996, sdujump, cloudseawang, changhe.tu\}@gmail.com,
a.g.cohn@leeds.ac.uk
}

\begin{document}

\maketitle

\input{01_abstract}

\input{02_introduction}

\input{03_related_works}

\input{04_method}

\input{05_experiment}

\input{06_conclusion}

\clearpage

\input{ijcai21.bbl}
\bibliographystyle{ijcai21}

\end{document}

%% file: 01_abstract.tex
\begin{abstract}
Scribble-supervised semantic segmentation has gained much attention recently for its promising performance without high-quality annotations. Due to the lack of supervision, confident and consistent predictions are usually hard to obtain. Typically, people handle these problems to either adopt an auxiliary task with the well-labeled dataset or incorporate the graphical model with additional requirements on scribble annotations. Instead, this work aims to achieve semantic segmentation by scribble annotations directly without extra information and other limitations. Specifically, we propose holistic operations, including minimizing entropy and a network embedded random walk on neural representation to reduce uncertainty. Given the probabilistic transition matrix of a random walk, we further train the network with self-supervision on its neural eigenspace to impose consistency on predictions between related images. Comprehensive experiments and ablation studies verify the proposed approach, which demonstrates superiority over others; it is even comparable to some full-label supervised ones and works well when scribbles are randomly shrunk or dropped.
\end{abstract}

%% file: 02_introduction.tex
\section{Introduction}
In recent years, the use of neural networks, especially convolutional neural networks, has dramatically improved semantic classification, detection, and segmentation~\cite{lecun2015deep}. As one of the most fine-grained ways to understand the scene, typically, semantic segmentation demands large-scale data with high-quality annotations to feed the network. However, the pixel-level annotating process for semantic segmentation is costly and tedious, limiting its flexibility and usability on some tasks that require rapid deployment~\cite{lin2016scribblesup}. As a consequence, the scribble annotations, which are more easily available, have become popular.

\begin{figure}[htbp]
\centering
\includegraphics[width=0.95\linewidth]{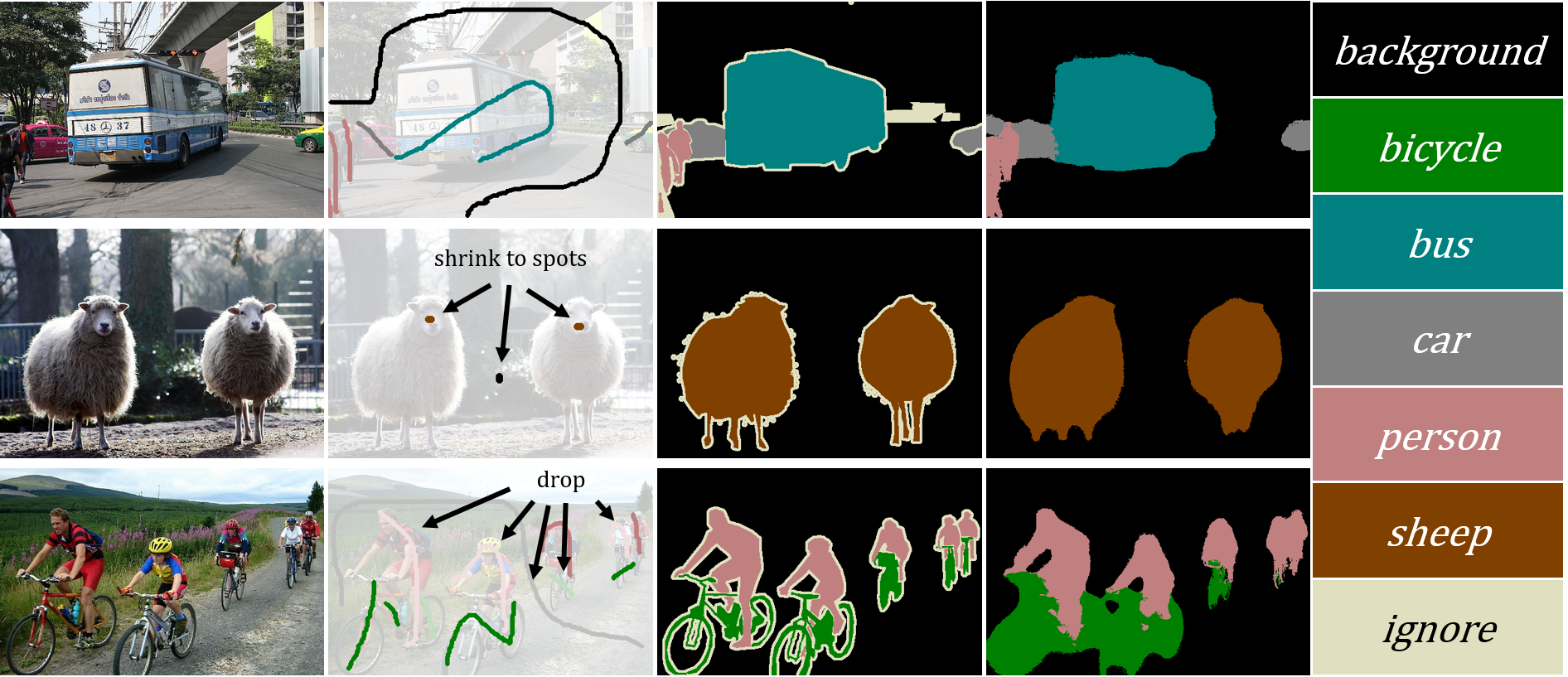}
\caption{From left to right: image, scribble annotation, ground truth, our prediction. From top to bottom: sample with regular, shrunk and dropped scribble annotation, respectively.}
\vspace{-3mm}
\label{fig:main}
\end{figure}

The main difficulty for scribble-supervised semantic segmentation lies in two aspects. (1) the scribble annotation is sparse and cannot provide enough supervision for the network to make confident predictions. (2) the scribble annotation varies from image to image, which is hard for the network to produce consistent results. As a consequence, \citeauthor{lin2016scribblesup} adopted the classic graphical model as post-processing to obtain the final dense predictions. Some works \cite{vernaza2017learning,wang2019boundary} turn to auxiliary task (edge detection) with well-labeled dataset for help,
so the heavy labor of annotation has not been relieved yet.
To avoid the post-processing and dependence on another well-labeled dataset,~\citeauthor{tang2018regularized} design a graphical model regularized loss to make predictions consistency within the appearance similar neighborhood, but did not consider semantic similarity. Moreover, they require every object in an image to be labeled, which is too strict for dataset preparation.

We address the task by a more flexible approach without introducing auxiliary supervision and constraints in this work. The approach can work properly when scribbles on some objects are randomly dropped or even shrunk to spots. Several representative results are listed in Fig.~\ref{fig:main}. We propose two creative solutions for the problems of confidence and consistency mentioned above. To reduce the uncertainty when supervision is lacking, we take advantage of two facts related to semantic segmentation. The first one is that each pixel only belongs to one category (deterministic), and there is only one channel of output neural representation that plays the dominant role. The second one is neural representations should be uniform within internal object regions. Accordingly, we develop the first solution on neural representation, including two specific operations, minimizing entropy to encourage deterministic predictions and a network embedded random walk module to promote uniform intermediates. Besides, the transition matrix of a random walk will also be useful for consistency enhancement later. In general, we make up for the lack of supervision with scribble annotations by taking advantage of two priors, deterministic and uniform.

We propose to adopt self-supervision during training as the second solution for inconsistent results caused by varying scribble annotations from image to image, which imposes consistency on the neural representation before and after certain input transformation~\cite{laine2016temporal}. However, consistency over the whole neural representation usually is not necessary for semantic segmentation, especially for regions belonging to the background category, which usually are semantically heterogeneous. When these regions are distorted and changed heavily after transformation, it is hard for the network to generate consistent output and may confuse the network in some scenarios. With that in mind and given the transition matrix of a random walk, we propose to set self-supervision on the main parts of images by imposing consistent loss on the eigenspace of transition matrix. The idea is inspired by spectral methods~\cite{von2007tutorial}, who observed that eigenvectors of a Laplacian matrix have the capability to distinguish the main parts in images, and some methods use this property for clustering~\cite{ng2002spectral} and saliency detection~\cite{jiang2019super}. Since the eigenspace of a transition matrix has a close relation to the one of a Laplacian matrix, our self-supervision on transition matrix's eigenspace will also focus on the main image parts.



The proposed approach demonstrates consistent superiority over others on the common scribble dataset and is even comparable to some fully supervised ones. Moreover, we further conduct experiments when scribbles are gradually shrunk and dropped. The proposed approach can still work reasonably, even the scribble shrunk to a spot or dropped significantly. Careful ablation studies are made to verify the effectiveness of every operation. Finally, in supplementary material, the code and dataset are open-sourced.

%% file: 03_related_works.tex
\section{Related Work}
Scribble-supervised semantic segmentation aims to produce dense predictions given only sparse scribbles. Existing deep learning-based works can usually be divided into two groups: 1) Two-stage approaches~\cite{lin2016scribblesup,vernaza2017learning}, which first obtain full mask pseudo-labels by manipulating scribble annotations, and then train the network as usual using semantic segmentation with pseudo-labels.
2) Single-stage approaches~\cite{tang2018normalized,tang2018regularized}, which directly train the network using scribble annotations by a specific loss function and network structure. While two-stage approaches can be formulated as regular semantic segmentation, single-stage approaches are usually defined to minimize the following function:
\begin{equation}
    L=\sum_{p\in\Omega_{\mathcal{L}}}c(s(x)_p,y_p)+\lambda\sum_{p,q\in\Omega}u(s(x)_p,s(x)_q),
\label{eq:eq1}
\end{equation}
where $\Omega$ is point (pixel) set, $\Omega_{\mathcal{L}}$ is point set with scribble annotations, $s(x)_i$ represents prediction at point $i$ given input $x$, and $y_i$ is the corresponding ground truth. The first term measures the error with scribble annotations and usually is in the form of cross-entropy. The second term is a pair-wise regularization to help generate uniform predictions. The two terms are harmonized by a weight parameter $\lambda$.

For scribble-supervised semantic segmentation, the graphical model has been prevalently adopted in either two-stage approaches for generating pseudo-label or one-stage approaches for loss design. \citeauthor{lin2016scribblesup} iteratively conduct label refinement and network optimization through a graphical model. \citeauthor{vernaza2017learning} generate high-quality pseudo-labels for full-label supervised semantic segmentation by optimizing graphical model with edge detector learned from another well-labeled dataset. These two works require iterative optimization or an auxiliary dataset. Instead, \citeauthor{tang2018regularized} add soft graphical model regularization into the loss function and explicitly avoid graphical model optimization. Besides, some only work well on a dataset where every object is labeled by at least one scribble. In general, most existing works have not provided a flexible and efficient solution to scribble-supervised semantic segmentation yet.



%% file: 04_method.tex
\section{Method}\label{sec:method}
Scribble-supervised semantic segmentation usually suffers from uncertain and inconsistent predictions due to lack of supervision and varying annotations from image to image. In this work, we propose two solutions, \emph{viz.} uncertainty reduction on neural representation and self-supervision on neural eigenspace to address these problems. Compared with others, we do not rely on auxiliary tasks with well-labeled datasets and additional requirements for annotation preparation.

\begin{figure*}[t]
\centering
\includegraphics[width=0.8\linewidth]{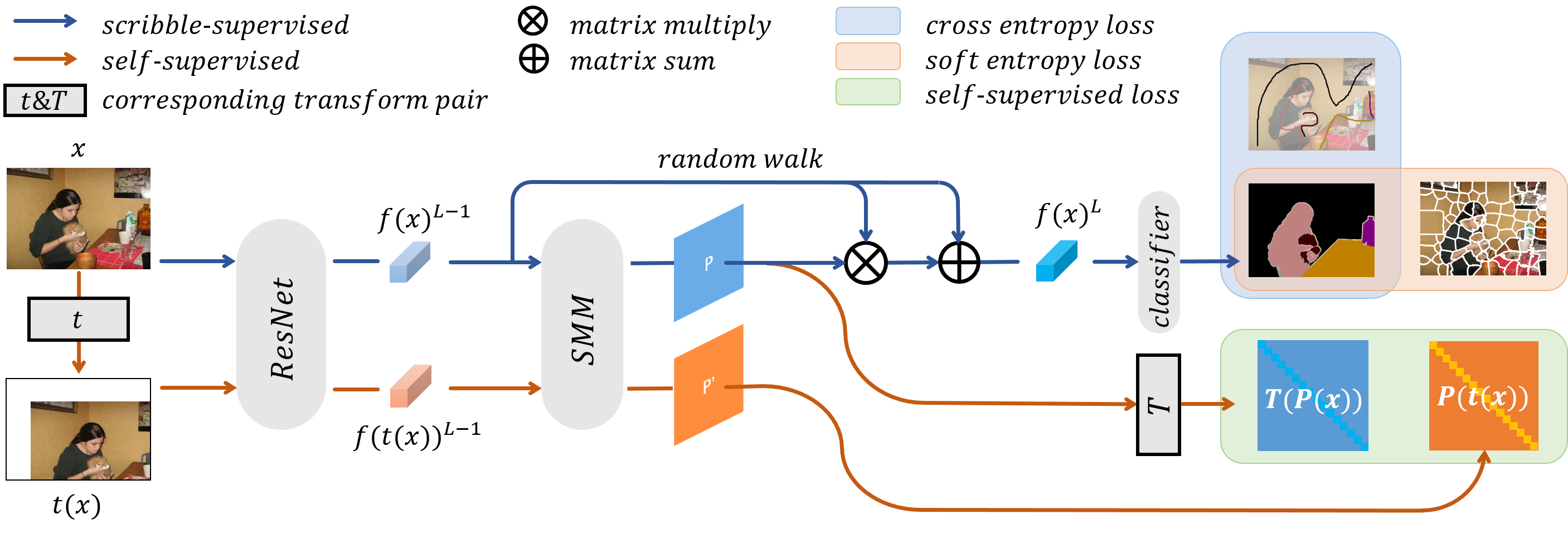}
\caption{Network Pipeline. We use blue and orange flows to represent scribble-supervised training and self-supervised training, respectively. Given an image and its transform, we pass them to ResNet to extract neural representations $f(x)^{L-1}$, from which similarity measurement module (SMM) computes transition matrix. A random walk is then carried out on $f(x)^{L-1}$. The results $f(x)^{L}$ are used for classification. Simultaneously, soft entropy by pseudo-boundaries is minimized to reduce uncertainty of neural representation, and self-supervised loss is set between transition matrices to realize self-supervision on neural eigenspace. During inference, only the blue flow is activated.}
\label{fig:pipeline}
\vspace{-3mm}
\end{figure*}

\subsection{Uncertainty Reduction on Neural Representation}\label{sec:uncertain}
To reduce the uncertainty on neural representation, we take advantage of priors that neural presentations should be deterministic and uniform for each semantic object. Thereby, holistic operations are developed and imposed on neural representation, including minimizing entropy and network embedded random walk.

\subsubsection{Minimizing entropy}
The entropy on neural representation is defined as:
\begin{equation}
\begin{aligned}
    E_{\Omega}=-\frac{1}{HW} \sum_{(i,j)\in\Omega}\sum_c{s(x)_{i,j,c}\cdot log(s(x)_{i,j,c}}),
\end{aligned}
\end{equation}
where $s(x)$ represents the prediction given the input $x$, and is of the size [\emph{H,W,C}] ($C$ is the number of categories). $s(x)_{i,j,c}$ represents the probability that the pixel at position $(i,j)$ belongs to the $c$-th category.

Entropy indicates the randomness of a system. According to a classical thermodynamic principle: \emph{Minimizing entropy results in minimum randomness of a system.} Thus, minimizing entropy on neural representation will reduce the uncertainty and force the network to produce deterministic predictions. However, uncertain predictions are inevitable in places such as object boundaries, and undifferentiated entropy minimization will cause network training conflict. Correspondingly, we propose to minimize entropy on neural representation excluding positions corresponding to object boundaries, leading to soft entropy:
\begin{equation}
\begin{aligned}
    E_{\Omega_{-\mathcal{B}}}=-\frac{1}{HW} \sum_{(i,j)\in\Omega_{-\mathcal{B}}}\sum_c{s(x)_{i,j,c}\cdot log(s(x)_{i,j,c}})
\end{aligned}
\end{equation}
where $\Omega_{-\mathcal{B}}$ is the point set that excludes object boundaries. In this way, minimizing soft entropy will reduce uncertainty and avoid potential conflicts on object boundaries. Since accurate boundaries are hard to acquire, we only use pseudo boundaries by the no-learning-based superpixel method, SLIC~\cite{achanta2012slic}. We note that entropy has been explored for some vision tasks, such as object detection~\cite{wan2018min}, but with different motivation and implementation. To our best knowledge, minimizing entropy is adopted to scribble-supervised semantic segmentation for uncertainty reduction for the first time.


\begin{figure}[t]
\centering
\includegraphics[width=0.45\textwidth]{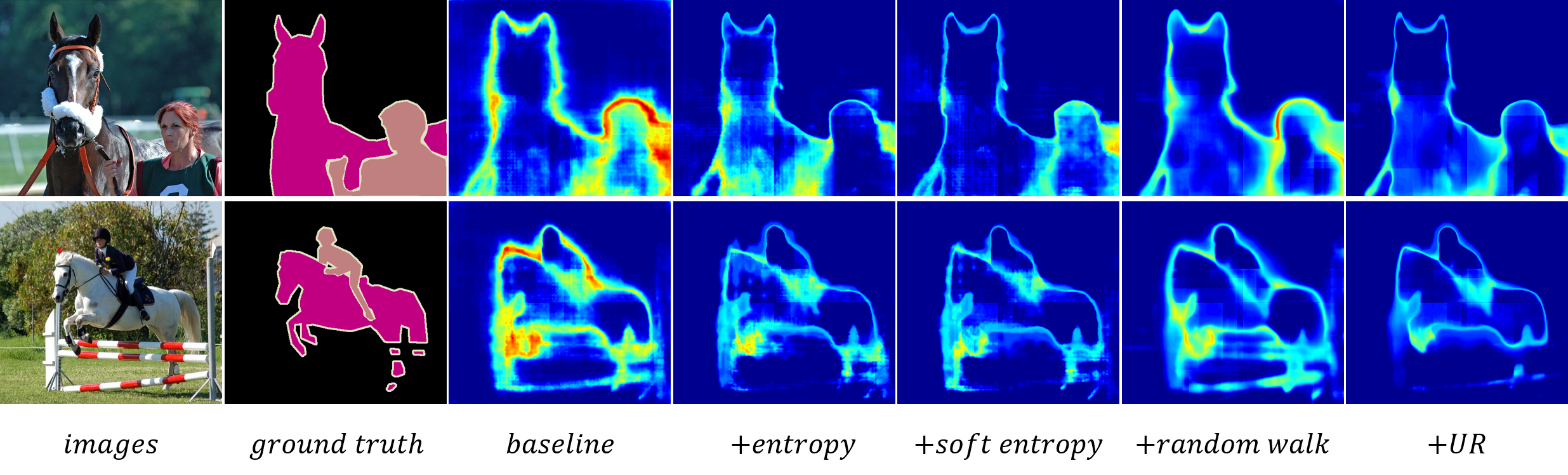}
\caption{Entropy map. Colder color indicates smaller entropy.}
\label{fig:entropy}
\vspace{-3mm}
\end{figure}

\subsubsection{Network embedded random walk}
A random walk operation is defined as:
\begin{equation}
    z=\alpha{Py}+(1-\alpha)y,
\end{equation}
$y$ is the initial state vector, $\alpha$ is a parameter that controls the degree of random walk, $P$ is transition matrix that measures the transition possibilities between every two positions, and we usually set similar positions with a large possibility of transition~\cite{von2007tutorial}. By definition above, the output state $z$ after random walk will have more similar states for similar positions, resulting in the uniform state within similar semantic/appearance regions.

Inspired by the characteristic of a random walk, we propose to embed this operation into the network for the uniform neural representation,
\begin{equation}
    f(x)^{L}=\alpha{Pf(x)^{L-1}}+f(x)^{L-1},
\label{eq:eq2}
\end{equation}
where $f(x)^{L-1}$ is the neural representation of input $x$ in layer $L$-$1$ and $f(x)^{L}$ is the neural representation after random walk in layer $L$, they are both of dimension [\emph{M,N,K}]. We set $\alpha$ as a learnable parameter to be trained during training and define the probabilistic transition matrix $P$ as:
\begin{equation}
    P=softmax({f(x)^{L-1}}^Tf(x)^{L-1}),
\label{eq:eq4}
\end{equation}
where $f(x)^{L-1}$ is flattened to [\emph{MN,K}], and ${f(x)^{L-1}}^Tf(x)^{L-1}$ will produce a matrix of dimension [\emph{MN,MN}]. By $softmax$ in the horizontal direction, we generate a suitable probabilistic transition matrix $P$ with all units are positive and every row of the matrix is summed to $1$.

A random walk has been frequently used for semantic segmentation tasks~\cite{bertasius2017convolutional,jiang2018difnet,ahn2018learning,Araslanov_2020_CVPR}. However, most of them use a random walk to diffuse the pseudo-label or refine the initial predictions. Instead, we use a random walk on neural representation for uniform and uncertainty reduction when given only scribble annotations.

\subsubsection{Uncertainty reduction verification}
In this part, we verify how the uncertainty is reduced by the two operations mentioned above. Given several randomly selected samples, we measure pixel-level entropy maps for predictions obtained by \emph{baseline}, networks with proposed operations individually (\emph{+entropy},\emph{+soft entropy},\emph{+random walk}) and together (\emph{+uncertainty reduction (UR)}). The results are visualized in Fig.~\ref{fig:entropy}. As expected, the entropy is decreased by the proposed two operations, and using them both leads to minimum entropy, albeit object boundaries remaining uncertain. The detailed setting of networks will be stated later.

\begin{figure}[htbp]
\centering
\includegraphics[width=0.40\textwidth]{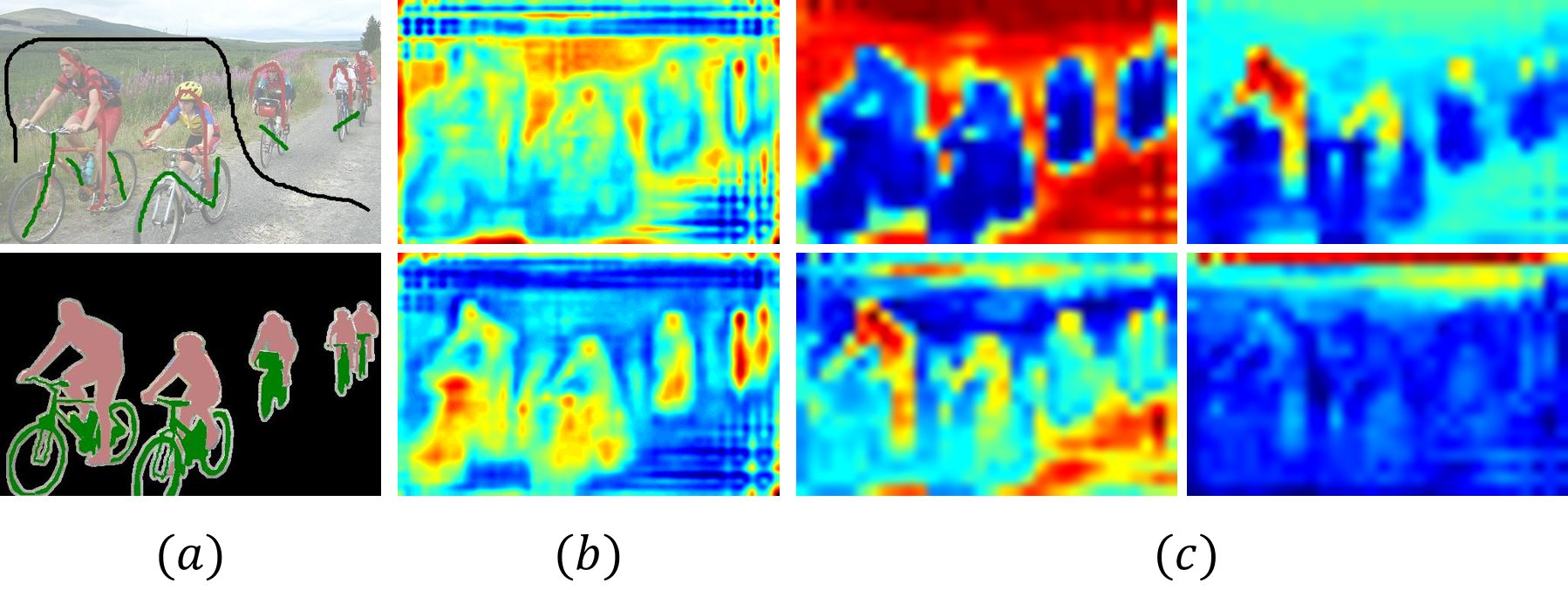}
\caption{(a) From top to bottom: scribble-annotation and ground-truth. (b) From top to bottom: Neural representation before and after a random walk. (c) Leading eigenvectors of the transition matrix.}
\label{fig:eigenvector}
\vspace{-3mm}
\end{figure}



\subsection{Self-Supervision on Neural Eigenspace}~\label{sec:selfsupervision}
Self-supervision computes the misfit between the network's intermediates of the input and its transform, which forces the network to produce consistent outputs. Self-supervision has been utilized for unsupervised learning tasks to provide unsupervised loss~\cite{laine2016temporal,mittal2019semi}.

We adopt self-supervision to address the issue of inconsistent results caused by varying scribble annotations from image to image. Several issues need to be considered when applying self-supervision loss in this work: (1) where the self-supervision is involved; (2) how the self-supervision loss is calculated; (3) what kinds of the transform will be used. We address these issues in the following.

\subsubsection{Self-supervision}
The most straight forward way to implement self-supervision is to compute the difference between neural representations of the input and its transform:
\begin{equation}
    ss(x,\phi)=l(T_\phi(f(x)),f(t_\phi(x))),
\label{eq:eq3}
\end{equation}
where $t_\phi$ denotes the transform operation on $x$ with $\phi$ parameter, while $T_\phi$ corresponds to the transform operation on $f(x)$ ($t_\phi$ and $T_\phi$ are a pair of corresponding transforms for self-supervision). $l$ is the metric to measure difference.
We denote this kind of consistency as $ss(x,\phi)$. 
Given operations in Sec.~\ref{sec:uncertain}, There are several obvious places to apply self-supervision, \emph{e.g.} neural representations $f(x)^{L-1}$ and $f(x)^{L}$.


\subsubsection{Self-supervision on neural eigenspace}
However, as for the semantic segmentation task with self-supervision, we argue that directly calculating loss on the whole neural representation is not necessary and may not be optimal. When the image is distorted heavily after the transform, some parts of its neural representation will change greatly, so minimizing Eq.~\ref{eq:eq3} will be hard and even ambiguous. In this work, given the transition matrix $P$ of random walk in Sec.~\ref{sec:uncertain}, we propose to apply self-supervision on the neural eigenspace of $P$.

The eigenspace of transition matrix $P$ and that of the normalized Laplacian matrix $L$ have close relationships~\cite{von2007tutorial}. It can be proved that $\Lambda_P$=$1-\Lambda_L$ and $U_P$=$U_L$ ($\Lambda$ denotes a diagonal matrix with eigenvalues as entries, $U$ denotes a matrix with eigenvectors as columns). According to \cite{Jiang_2015_ICCV,jiang2019super}, columns of $U_L$ have the capability to distinguish the main parts of the images. So, $U_P$ will also inherit this property. We visualize several eigenvectors of $P$ in Fig.~\ref{fig:eigenvector}. As can be seen, compared with the original neural representations $f(x)^{L-1}$ and $f(x)^{L}$, the eigenvectors of $P$ are better able to distinguish the main parts from others and neglect some details, though $P$ is also computed from neural representation. Based on the above analysis, we define the self-supervision as,
\begin{equation}
\begin{aligned}
    ss(x,\phi)&=l(T_\phi(U_P(x)),U_P(t_\phi(x)))\\
    &+l(T_\phi(\Lambda_P(x)),\Lambda_P(t_\phi(x))).
\end{aligned}
\label{eq:eq7}
\end{equation}

\subsubsection{Soft eigenspace self-supervision}
Eq.~\ref{eq:eq7} requires explicit eigendecomposition, which is time-consuming, especially within the deep neural network context. Though there are some approximation methods~\cite{dang2018eigendecomposition,wang2019backpropagation,sun2019neural} proposed, their efficiency and stability are still far from satisfactory. To this end, we develop soft eigenspace self-supervision, which avoids explicit eigendecomposition. Firstly, in view of the fact that the matrix's trace is equal to the sum of its eigenvalues, we measure the consistency on $\Lambda$ by computing the difference to the trace of $P$, $tr(P)$. Secondly, given the consistency on the $\Lambda$, we propose to measure the consistency on the $P$ to obtain consistent $U$ indirectly. In other words, the soft eigenspace self-supervision loss is defined as:
\begin{equation}
\begin{aligned}
    ss_P(x,\phi)&=l_1(T_\phi(P(x)),P(t_\phi(x)))\\
    &+\gamma\ast l_2(T_\phi(tr(P(x))),tr(P(t_\phi(x)))),
\end{aligned}
\label{eq:eq8}
\end{equation}
where $P(x)$ denotes $P$ for input $x$, $tr(P(x))$ is the trace of $P(x)$. Since $P(x)$ is a probabilistic transition matrix, we define $l_1$ as Kullback-Leibler Divergence, and use the $L_2$ norm for $l_2$. $\gamma$ is the weight to control the two terms.

\subsubsection{Computing efficiency}
We set two linear transform operations for self-supervision, horizontal flip, and translation, $\phi\in$(horizontal flip, translation). Compared with the transform that impacts neural representation, any transform will lead to a complex change on $P$ and complicate the computation. However, since all transform operations are linear, the probabilistic transition matrix after transform can be expressed as the multiplication of the original $P$ with predefined computing matrices to facilitate Eq.~\ref{eq:eq8} computation. $T_{\phi}(P(x))$ can be defined as:
\begin{equation}
\begin{aligned}
    T_{\phi}(P(x))=T_{\phi}r\cdot{P(x)}\cdot{T_{\phi}c},
\end{aligned}
\end{equation}
where $T_{\phi}r$ and $T_{\phi}c$ are predefined computing matrices for transform $\phi$. 
Details are in the supplementary material.


%% file: 05_experiment.tex
\begin{table}[htbp]
    \centering
    \setlength{\belowcaptionskip}{3pt}
    \begin{tabular}{c|c|c|c}
    \toprule[1pt]
        \multicolumn{3}{c|}{Uncertainty Reduction}&\multirow{2}{*}{mIoU}\\
        \cline{1-3}
        entropy&boundary&random walk&~\\
        \hline
        \hline
         
        ~&~&~&66.8\\
         
        ~&~&\checkmark&69.3\\
         
        \checkmark&~&~&69.4\\
         
        \checkmark&\checkmark&~&70.0\\
         
        \checkmark&\checkmark&\checkmark&\textbf{70.9}\\
    \bottomrule[1pt]
    \end{tabular}
    \caption{Ablation study for uncertainty reduction.}
    \vspace{-3mm}
    \label{tab:ablation_ur}
    \vspace{-3mm}
\end{table}

\begin{table}[htbp]
    \centering
    \setlength{\belowcaptionskip}{3pt}
    \begin{tabular}{c|c|c}
    \toprule[1pt]
        \multicolumn{2}{c|}{Self-Supervision}&\multirow{2}{*}{mIoU}\\
        \cline{1-2}
        transform operation&location&~\\
        \hline
        \hline
        \multirow{3}{*}{flip}&$f(x)^{L-1}$&72.5\\

        ~&$f(x)^L$&72.5\\

        ~&$Eigenspace$&\textbf{72.9}\\
        \hline
        \multirow{3}{*}{translation}&$f(x)^{L-1}$&71.9\\

        ~&$f(x)^L$&70.0\\
        
        ~&$Eigenspace$&\textbf{72.6}\\
        \hline
        \multirow{3}{*}{random}&$f(x)^{L-1}$&72.4\\

        ~&$f(x)^L$&71.6\\
        
        ~&$Eigenspace$&\textbf{73.0}\\
    \bottomrule[1pt]
    \end{tabular}
    \caption{Ablation study for self-supervision.}
    \label{tab:ablation_ss}
    \vspace{-3mm}
\end{table}

\section{Implementation}
The network is illustrated in Fig.~\ref{fig:pipeline}, including two modules (ResNet to extract features, and Similarity Measurement Module (\emph{SMM}) to compute probabilistic transition matrix), one specific process (random walk), and three loss functions (soft entropy, self-supervision, and cross-entropy). These components realize uncertainty reduction on neural representation and self-supervision on neural eigenspace.

A random walk is embedded in the network's computation flow and conducted on the final layer right before the classifier, strictly following Eq.~\ref{eq:eq2} with learned $\alpha$ that controls the degree of random walk. Similarity Measurement Module computes the inner product distance between any pairs of neural representation elements and forms the probabilistic transition matrix $P$ as Eq.~\ref{eq:eq4}.

We use the pre-trained ResNet~\cite{he2016deep} with dilation~\cite{chen2017deeplab} as the backbone to extract initial neural representations. The total loss in our work is defined as:
\begin{equation}
\begin{aligned}
L=\sum_{p\in\Omega_{\mathcal{L}}}c(s(x)_p,y_p)+\omega_1 E_{\Omega_{\mathcal{-B}}}+\omega_2\ast ss_P(x,\phi),
\end{aligned}
\label{eq:total}
\end{equation}
where $\omega_1$ and $\omega_2$ are predefined weights. The first term measures the divergence of prediction to ground truth at positions with scribbles. The second term computes entropy within regions excluding pseudo-boundaries. The third term is self-supervision on eigenspace. By minimizing Eq.~\ref{eq:total}, we are training the network to approximate scribbles when they are available, produce confident predictions and consistent outcome eigenspace. The random walk embedded in the network will also help generate uniform intermediates. Consequently, we overcome difficulties of scribble-supervised semantic segmentation when given sparse and random annotations.

The training process has two steps. In the beginning, only the first two terms participate. At this time, the network may not perform well initially, and self-supervision will not bring benefits but prevent the optimization. After the network gets reasonable performance, the whole Eq.~\ref{eq:total} is activated.

\begin{figure}[htbp]
\centering
\includegraphics[width=0.8\linewidth]{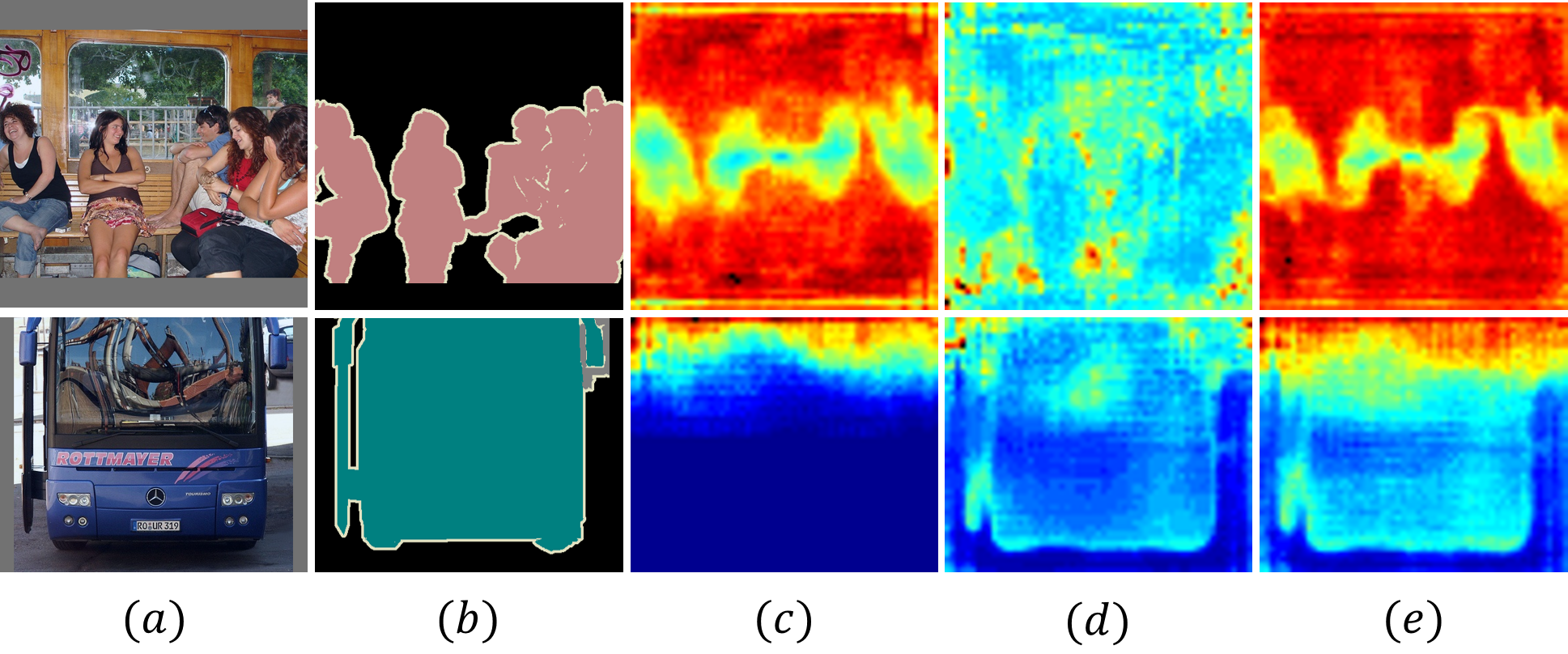}
\caption{Variation by different self-supervision. (a) input image, (b) ground truth, (c) (d) (e) variation on $f(x)^{L-1}$, $P(x)$ and $f(x)^L$, respectively. The first row shows variations for the flip operation, while the second row is for the translation operation.}
\label{fig:variations}
\vspace{-3mm}
\end{figure}

\begin{table}[htbp]
    \centering
    \setlength{\belowcaptionskip}{3pt}
    \begin{tabular}{c|ccc}
    \toprule[1pt]
    &$f(x)^{L-1}$&$f(x)^L$&$P(x)$\\
    \hline
    \hline
    flip&52.8\%&37.5\%&6.7\%\\
    
    translation&7.5\%&12.0\%&3.5\%\\
    \bottomrule[1pt]
    \end{tabular}
    \caption{Variation comparison under the same transform operation.}
    \label{tab:ablation_background}
    \vspace{-3mm}
\end{table}

\section{Experiment}
\subsection{Experiment Setting}
\subsubsection{Datasets}\label{sec:dataset}
We make comparison on the common scribble-annotated dataset, \emph{scribblesup} ~\cite{lin2016scribblesup}. This dataset has 21 classes (including an ignore category) with every object in the image labeled by at least one scribble. However, our approach can work without preconditions. 
To verify our advantages, we further prepare two variants of \emph{scribblesup} with the same training and validation partition. The first one is \emph{scribble-drop}, where every object in image randomly drops (\emph{i.e.} deletes) all scribbles. The second one is \emph{scribble-shrink}, where every scribble in the image is shrunk randomly (even to a spot). We test many settings of the drop and shrink rate. 

\subsubsection{Compared methods}
We compare with recently proposed scribble-supervised methods including scribblesup~\cite{lin2016scribblesup}, RAWKS~\cite{vernaza2017learning}, NCL~\cite{tang2018normalized}, KCL~\cite{tang2018regularized}, BPG-PRN~\cite{wang2019boundary}, and also point supervised methods (What’sPoint~\cite{bearman2016s}).
Besides, the full-label supervised method (DeepLabV2~\cite{chen2017deeplab}) is also compared. We use mIoU as the main metric for evaluation. When comparing with others, we refer to their reported scores if available.

\subsubsection{Hyper-parameters}
All training images are randomly scaled (0.5 to 2), rotated (-10 to 10), blurred, and flipped for data augmentation, then cropped to [465,465] before feeding to the network. $f(x)^{L-1}$ and $f(x)^{L}$ are of spatial dimension [59,59]. All the computations are carried out on NVIDIA TITAN RTX GPUs. The supplementary material details the setting of $\gamma$, $\omega_1$ and $\omega_2$.


\begin{table}[htbp]
    \centering
    \setlength{\belowcaptionskip}{3pt}
    \scalebox{0.9}{
    \begin{tabular}{l|c|c|c|c}
    \toprule[1pt]
        Method&Ann.&Backbone&wo/ CRF&w/ CRF\\
        \hline
        \hline
        What'sPoint&$\mathcal{P}$&$VGG16$&46.0&-\\
        
        DeepLabV2&$\mathcal{F}$&$ResNet101$&76.4&77.7\\
        \hline
        scribblesup&$\mathcal{S}$&$VGG16$&-&63.1\\

        RAWKS&$\mathcal{S}$&$ResNet101$&59.5&61.4\\
        
        NCL&$\mathcal{S}$&$ResNet101$&72.8&74.5\\

        KCL&$\mathcal{S}$&$ResNet101$&73.0&75.0\\
        
        BPG-PRN&$\mathcal{S}$&$ResNet101$&71.4&-\\
        \hline
        
        ours-ResNet50&$\mathcal{S}$&$ResNet50$&73.0&74.7\\
        

        ours-ResNet101&$\mathcal{S}$&$ResNet101$&\textbf{74.4}&\textbf{76.1}\\
    \bottomrule[1pt]
    \end{tabular}}
    \caption{Performance on validation set of $Scribblesup$. The annotation type (Ann.) indicates: $\mathcal{P}$–point, $\mathcal{S}$–scribble and $\mathcal{F}$–full label.}
    \label{tab:all}
    \vspace{-3mm}
\end{table}

\begin{table*}[htbp]
    \centering
    \setlength{\belowcaptionskip}{3pt}
    \begin{tabular}{l|c|c|c|c|c||l|c|c|c|c}
    \toprule[1pt]
    drop rate&0.1&0.2&0.3&0.4&0.5&shrink rate&0.2&0.5&0.7&1\\
    \hline
    \hline
    baseline&66.3&65.4&65.1&63.9&63.8&baseline&66.2&64.8&63.8&57.0\\
    
    +Uncertainty Reduction&69.6&69.1&68.0&67.3&67.2&+Uncertainty Reduction&69.7&67.3&65.3&59.0\\
    
    +Self-Supervision&72.2&71.4&71.4&69.8&69.5&+Self-Supervision&72.1&70.4&68.8&62.8\\
    \bottomrule[1pt]
    \end{tabular}
    \caption{mIoU scores on \emph{scribble-drop} and \emph{scribble-shrink} with different drop and shrink rates.}
    \label{tab:drop_ratios}
    \vspace{-3mm}
\end{table*}

\subsection{Ablation Study}
In this part, we investigate all operations involved in Sec.~\ref{sec:method}. We use \emph{Scribblesup} dataset for training and validation. Firstly, we do an ablation study for operations in Sec.~\ref{sec:uncertain}. Starting with baseline (\emph{ResNet50}), we gradually add entropy minimization, boundary exclusion, and random walk, obtaining networks with different combinations. We report mIoU in Tab.~\ref{tab:ablation_ur}. The first row is the baseline. We can observe that all operations can obtain better performance, and using them all leads to the best performance. We get 4.1\% improvement by uncertainty reduction on neural representation in total.

Secondly, we do an ablation study for self-supervision in Sec.~\ref{sec:selfsupervision}. Starting with the network getting best score (70.9\%) in Tab.~\ref{tab:ablation_ur}, we compare networks after further training by self-supervision on $f(x)^{L-1}$, $f(x)^{L}$ and eigenspace of $P(x)$ with different transform operations. We report mIoU in Tab.~\ref{tab:ablation_ss}. We observe that not all of them would bring improvement, self-supervision on $f(x)^{L}$ with translation operation even deteriorates the performance. However, self-supervision on the eigenspace of $P(x)$ improves the performance consistently, and randomly selecting transform operations achieves the best performance. We get $2.1\%$ (from 70.9\% to 73.0\%) improvement by self-supervision on the neural eigenspace. 

The eigenspace of $P$ is the feature located behind $f(x)^{L-1}$ and in front of $f(x)^L$. To delve into the reason why self-supervision on eigenspace outperforms others, in Tab.~\ref{tab:ablation_background}, we show mean variations of $f(x)^{L-1}$, $f(x)^L$ and $P(x)$ under the same transform (no self-supervision applied yet). The variation is measured by the relative error.
As can be seen, the same transform will always lead to less variation on $P(x)$. In Fig.~\ref{fig:variations}, we visualize variation for self-supervision on different places. 
Compared with $f(x)^{L-1}$ and $f(x)^L$, variation on $P(x)$ is smaller in background regions.
This phenomenon indicates that self-supervision on $P(x)$ can focus on the main parts and avoid training conflicts on the background category with unstable semantics, relieving the training burden.

\subsection{Quantitative Results}
We get mIoU $73.0\%$ with ResNet50 and $74.4\%$ with ResNet101 on the \emph{Scribblesup} dataset. When comparing with others, we also report performance with CRF as others. Tab.~\ref{tab:all} lists all scores for compared methods. In addition to scribble-supervised methods, we also show methods with other label types, such as point and full-label. The proposed method reaches state-of-the-art performance compared with other scribble-supervised methods and is even comparable to the full-label one. The reported full-label method (DeepLabV2) was additionally pre-trained on COCO dataset. The supplementary material has more results by different label types.


It should be noted that some methods, such as~\cite{lin2016scribblesup,vernaza2017learning,tang2018regularized}, require every object is labeled. However, ours does not have this limit. In Tab.~\ref{tab:drop_ratios}, we show the performance under different drop and shrink rates on \emph{scribble-drop} and \emph{scribble-shrink} datasets. As can be seen, with our proposed solutions, we perform well when the drop rate and shrink rate increase, even when all scribbles were shrunk to spots. 



\begin{figure}[htbp]
\centering
\includegraphics[width=1\linewidth]{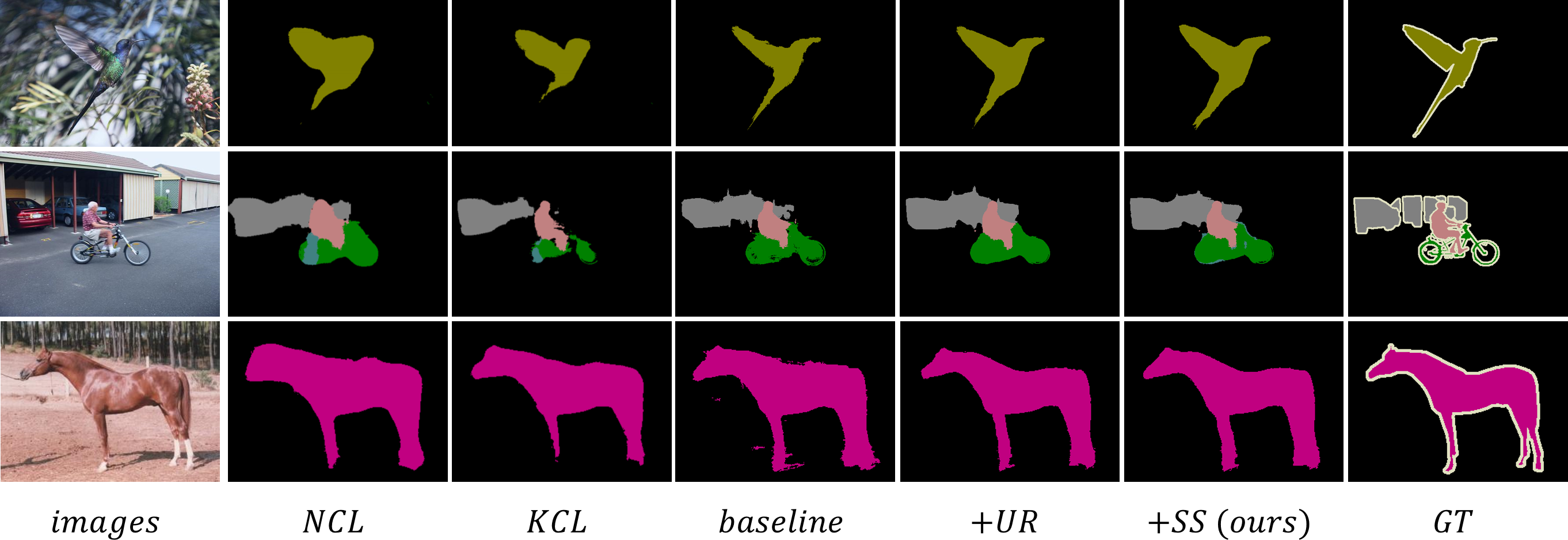}
\caption{Visual comparison between ours and others.}
\label{fig:VisualizationResults}
\vspace{-3mm}
\end{figure}

\subsection{Qualitative Results}
Fig.~\ref{fig:VisualizationResults} shows the visual comparison between NCL, KCL, and ours on three images from the test set of \emph{scribblesup}. With the proposed uncertainty reduction (UR) and self-supervision (SS), results are gradually refined and show significant promotion over the baseline and others.
(The results and scores in this section are all from the validation set. The supplementary material has more results on \emph{scribble-drop} and \emph{scribble-shrink} datasets.)


\begin{table}[htbp]
    \centering
    \setlength{\belowcaptionskip}{3pt}
    \begin{tabular}{l|c|c|c}
    \toprule[1pt]
    &P (M)&M (MB)&S (it/s)\\
    \hline
    \hline
    Baseline&41.72&2015.5&4.21\\
    
    +Uncertainty Reduction&+1.24&+37.2&+0.32\\
    
    +Self-Supervision&+0&+10.8&+0\\
    \bottomrule[1pt]
    \end{tabular}
    \caption{Parameters (P), memory (M), inference speed (S).}
    \label{tab:efficiency}
    \vspace{-3mm}
\end{table}


\subsection{Computational Resources Consumed}
The consumed resources are measured in Tab.~\ref{tab:efficiency}. Self-supervision does not need to preserve intermediates nor has extra parameters and is only adopted during training. Random walk only requires moderate space to save the transition matrix. Consequently, the cost of the proposed solutions is acceptable.


%% file: 06_conclusion.tex
\section{Conclusion}
In this work, we recognize that semantic segmentation given only scribble annotations will cause uncertain and inconsistent predictions. Accordingly, we develop two creative solutions, uncertainty reduction on neural representation to produce confident results, and self-supervision on neural eigenspace for consistency in output. No additional information and requirement for annotation preparation is needed.
Thorough ablation studies and intermediate visualization have verified the effectiveness of the proposed solutions. Finally, we reach state-of-the-art performance compared with others, even comparable to the full-label supervised ones. Moreover, the proposed approach still works when scribbles are randomly dropped or shrunk. 
